\documentclass[letterpaper]{article} % DO NOT CHANGE THIS
\usepackage{aaai2026}  % DO NOT CHANGE THIS
\usepackage{times}  % DO NOT CHANGE THIS
\usepackage{helvet}  % DO NOT CHANGE THIS
\usepackage{courier}  % DO NOT CHANGE THIS
\usepackage[hyphens]{url}  % DO NOT CHANGE THIS
\usepackage{graphicx} % DO NOT CHANGE THIS
\usepackage{natbib}  % DO NOT CHANGE THIS AND DO NOT ADD ANY OPTIONS TO IT
\usepackage{caption} % DO NOT CHANGE THIS AND DO NOT ADD ANY OPTIONS TO IT
\usepackage{algorithm}
\usepackage{algorithmic}

\usepackage{newfloat}
\usepackage{listings}
\DeclareCaptionStyle{ruled}{labelfont=normalfont,labelsep=colon,strut=off} % DO NOT CHANGE THIS
\floatstyle{ruled}
\newfloat{listing}{tb}{lst}{}
\floatname{listing}{Listing}
\usepackage{needspace} 
\usepackage{booktabs}
\usepackage{array}
\usepackage{enumitem}
\usepackage{pifont}
\usepackage{multicol}
\usepackage{fancyhdr}
\usepackage[utf8]{inputenc}
\usepackage[most]{tcolorbox}
\usepackage{pifont}

\usepackage{times}
\usepackage{helvet}
\usepackage{courier}
\usepackage{xcolor}

\title{Enhancing Long Video Question Answering with Scene-Localized Frame Grouping}

\author {
    Xuyi Yang\textsuperscript{\rm 1,3}\equalcontrib,
    Wenhao Zhang\textsuperscript{\rm 1,2}\equalcontrib
    Hongbo Jin\textsuperscript{\rm 2}\equalcontrib,
    Lin Liu\textsuperscript{\rm 3},
    Hongbo Xu\textsuperscript{\rm 1},
    Yongwei Nie\textsuperscript{\rm 4},
    Fei Yu\textsuperscript{\rm 1},
    Fei Ma\textsuperscript{\rm 1}\thanks{Corresponding author.}
}

\affiliations {
    \textsuperscript{\rm 1}Guangdong Laboratory of Artificial Intelligence and Digital Economy (SZ), Shenzhen, China\\
    \textsuperscript{\rm 2}School of Electronic and Computer Engineering, 
Peking University, Shenzhen, China\\
    \textsuperscript{\rm 3}School of Computer Science, Wuhan University, Wuhan, China\\
    \textsuperscript{\rm 4}University of Science and Technology of China, Hefei, China\\
    \textsuperscript{\rm 5}School of Computer Science an Engineering, South China University of Technology, Guangzhou, China\\
    xuyi-yang@whu.edu.cn,
    \{whzhang25, hbjin25\}@stu.pku.edu.cn, ll0825@mail.ustc.edu.cn,\\
    nieyongwei@scut.edu.cn, yufei@szu.edu.cn,
    \{xuhongbo, mafei\}@gml.ac.cn
}

\usepackage{bibentry}
\begin{document}
\maketitle

\begin{abstract}
Current Multimodal Large Language Models (MLLMs) often perform poorly in long video understanding, primarily due to resource limitations that prevent them from processing all video frames and their associated information. Efficiently extracting relevant information becomes a challenging task. Existing frameworks and evaluation tasks focus on identifying specific frames containing core objects from a large number of irrelevant frames, which does not align with the practical needs of real-world applications.
To address this issue, we propose a new scenario under the video question-answering task, SceneQA, which emphasizes scene-based detail perception and reasoning abilities. And we develop the LVSQA dataset to support the SceneQA task, which is built upon carefully selected videos from LVBench and contains a new collection of question-answer pairs to promote a more fair evaluation of MLLMs' scene perception abilities in long videos.
Inspired by human cognition, we introduce a novel method called SLFG. The core idea of SLFG is to combine individual frames into semantically coherent scene frames. By leveraging scene localization methods and dynamic frame reassembly mechanisms, SLFG significantly enhances the understanding capabilities of existing MLLMs in long videos. SLFG requires no modification to the original model architecture and boasts excellent plug-and-play usability. Experimental results show that this method performs exceptionally well in several long video benchmark tests. Code and dataset will be released at http://www.slfg.pkuzwh.cn.
\end{abstract}
\begin{figure}[htbp]
\centering
\includegraphics[width=0.39\textwidth]{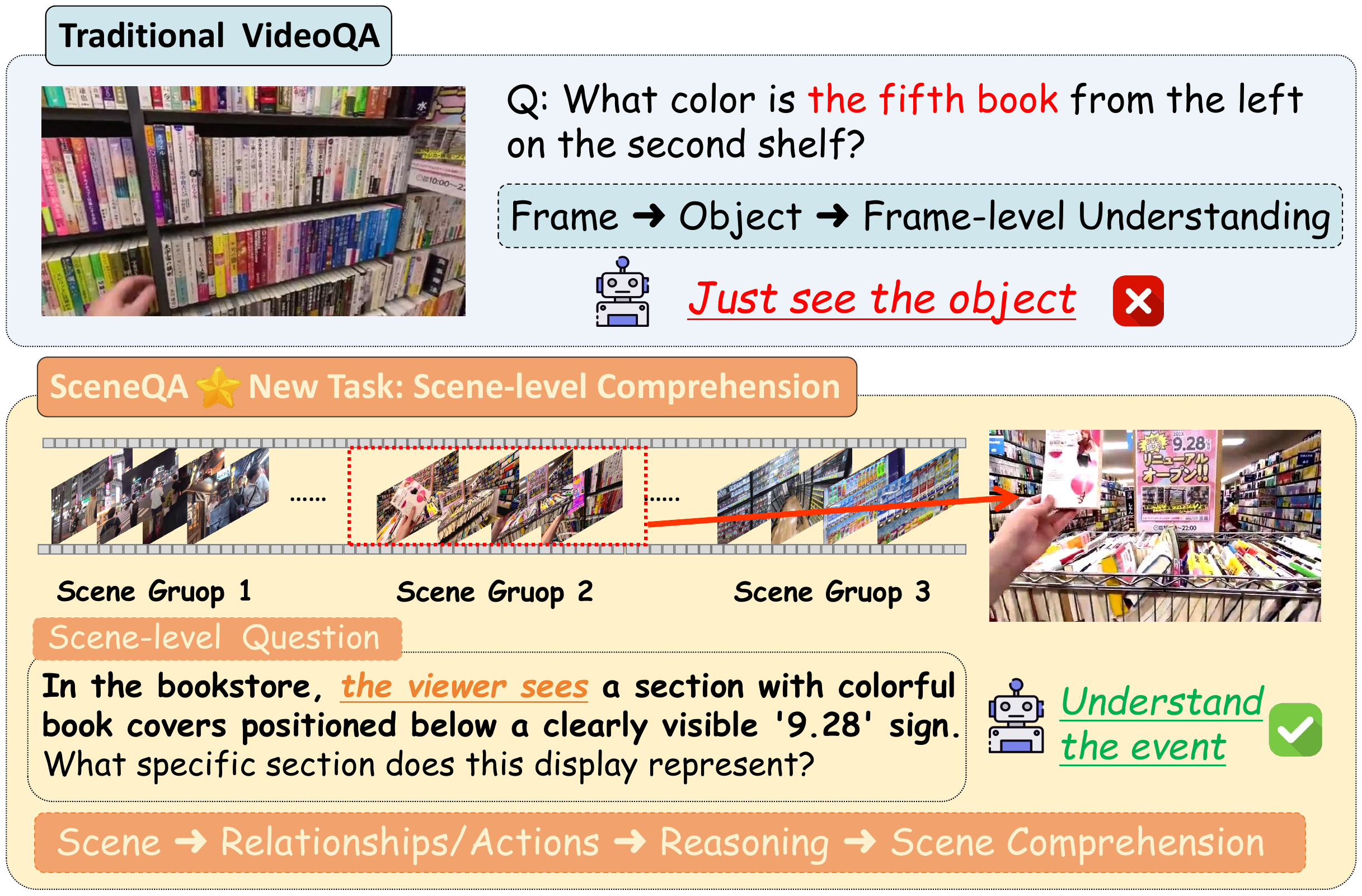}
\caption{An illustration of SceneQA task. Traditional VideoQA typically evaluates frame-level understanding by asking simple, direct questions. Our SceneQA task challenges models to comprehend coherent scenes as a whole, requiring them to answer complex questions that involve contextual reasoning about the relationships between multiple elements within the scene.
} \label{fig1}
\end{figure}
\section{Introduction}
% Multimodal Large Language Models (MLLMs) \cite{liu2023visual,cheng2024videollama,openai_gpt4o_2024,liu2024ppllava} have shown strong capabilities by aligning visual and textual information, enabling them to understand complex visual inputs such as videos. With stronger temporal dynamics and higher information density, videos have become central in multimodal research \cite{tang2023video,liang2024survey}. Video Question Answering (VideoQA) is widely used to evaluate models' temporal and visual-language reasoning abilities \cite{dai2025towards,chang2024survey}. Recent models like LLaVA-Video \cite{li2024llava,zhang2024video} and InternVL \cite{chen2024internvl} have advanced this area by using high-quality video-language data and temporal modeling. Nevertheless, effectively handling long videos remains a key challenge for existing MLLMs.

% Current studies and benchmarks for long videos \cite{wu2024longvideobench,zhou2024mlvu,wang2024LVBench} often focus on identifying specific frames containing core objects from a large number of irrelevant ones. We argue that this evaluation paradigm does not align with the practical needs of real-world applications, which demand perception of details and events within a broader scene. This promble highlights a critical gap: existing VQA tasks oversimplify the problem, neglecting true scene-level comprehension. 
Multimodal Large Language Models (MLLMs)  \cite{liu2023visual,cheng2024videollama,openai_gpt4o_2024,liu2024ppllava} have shown strong capabilities in understanding complex visual inputs like videos, with Video Question Answering (VideoQA)  \cite{dai2025towards,chang2024survey} serving as a key evaluation method. Despite advances from models like LLaVA-Video  \cite{li2024llava,zhang2024video} and InternVL  \cite{chen2024internvl}, effectively handling long videos remains a key challenge. Current studies and benchmarks  \cite{wu2024longvideobench,zhou2024mlvu,wang2024LVBench} often compound this issue by focusing on identifying specific frames with core objects from a large number of irrelevant ones. We argue that this evaluation paradigm does not align with real-world applications, which demand perception of details and events within a broader scene. This highlights a critical gap: existing VQA tasks oversimplify the problem, neglecting true scene-level comprehension.

To develop more general and efficient solutions, we reconsider how humans process long videos. Rather than analyzing each individual frame, humans typically skim through videos, remaining sensitive to scene transitions \cite{wang2023long}. Once a relevant scene is detected, they focus on examining the details within that specific segment. Drawing inspiration from this cognitive process, we argue that for models to truly understand long videos, the fundamental unit of analysis should shift from individual frames to semantically coherent scenes.

Building on this idea, we introduce a new task—SceneQA, a scene-localized long video question-answering task. SceneQA is designed to assess a model’s ability to accurately identify and focus on key scenes related to a given question within lengthy videos, thus placing higher demands on the model's ability to perceive details at the scene level.

To effectively tackle the challenges posed by SceneQA, we further propose \textbf{SLFG} 
 (Scene-Localized Frame Grouping), which aims to reduce frame redundancy at a higher semantic level by filtering frames through scene localization.
First, by performing dense sampling of the video and aggregating consecutive frames into fixed-granularity frame groups, we use a large language model to generate scene-level semantic representations, effectively compressing the input while retaining key visual information. Next, we extract the scene descriptions from the question and compute its semantic similarity with each frame group's scene descriptions, achieving precise alignment between the question and relevant content. Finally, we introduce a dynamic frame group reorganization module that adjusts the frame group structure based on similarity scores, merging high-relevance groups, discarding low-relevance groups, and expanding the time window of key segments to enhance information density and reasoning effectiveness within the limited context. The overall approach requires no modifications to the original model architecture, offering excellent plug-and-play compatibility and generalizability, making it suitable for enhancing the performance of existing MLLMs in long video understanding tasks.

To support the study and evaluation of SceneQA, we developed a dataset: \textbf{LVSQA} (Long Video Scene-level Question Answering Dataset). We carefully select 100 long videos (each over 30 minutes) from LVBench \cite{wang2024LVBench}, filter out those heavily reliant on subtitles, and have human experts redesign and expand the tasks from a purely visual perspective. We construct 500 high-quality question-answer pairs through a collaborative pipeline that combines MLLM-assisted generation with extensive human refinement. The question-answer pairs focus on detailed visual understanding, providing a challenging and targeted testbed for assessing models' scene-level localization and fine-grained visual reasoning in long videos.

Since our method is pluggable, we choose LLaVA-OneVision\cite{li2024llava} and LLaVA-Video\cite{zhang2024video} as the base models. After applying our method, both models achieve significant accuracy improvements across various evaluation sets. Specifically, on VideoMME$_{w/o~~long}$\cite{fu2024video}, our method elevates the performance of LLaVA-OneVision from \textbf{43.6\%} to \textbf{46.9\%}, and LLaVA-Video from \textbf{47.6\%} to \textbf{49.5\%}. On LVBench, LLaVA-OneVision with SLFG increases by \textbf{9.8\%} relatively. LLaVA-Video with SLFG increases by \textbf{8.4\%} relatively. At the same time, we also evaluate mainstream MLLMs on the LVSQA, and significant improvements are achieved after incorporating our method.
We believe that the experimental results reflect that a well-designed task can better assess a MLLM's ability to capture detailed visual information in long videos, laying a foundation for future exploration of various strategies in the field of long video understanding.

Our contributions can be summarized as follows:
\begin{itemize}[left=0pt,topsep=2pt] 
  \item We propose a new long video question answering task—SceneQA, which emphasizes scene-localized video understanding, and construct a dedicated benchmark dataset LVSQA to support this task, to promote research on long video understanding.
  \item We propose SLFG (Scene-Localized Frame Grouping), which aggregates frames by grouping them and understands key information in long videos at a higher-level scene semantic space.
  \item Experiments on mainstream long video benchmarks demonstrate that our method can significantly enhance the ability of existing MLLMs to perceive long video details. 
  
\end{itemize}
% Frustratingly, during the evaluation of the methods, we identified several issues with existing datasets. On one hand, some videos heavily rely on subtitle information\cite{zohar2024apollo}, making it difficult to understand the content based solely on visual cues. On the other hand, some questions are too broad, interfering with the model's ability to focus and understand the key visual aspects.
% To address this issue, we developed the LVSQA\textbf{(Long Video Scene-level Question Answering)}. We carefully selected 100 long videos from LVBench\cite{wang2024LVBench}, each with a duration of over 30 minutes. Human experts enhanced and expanded the original tasks and meticulously organized 500 question-answer pairs. 
% Unlike existing benchmarks, the tasks of LVSQA are carefully designed to focus on detailed questions within the scope of visual understanding. Additionally, the carefully selected videos eliminate the interference of subtitle information, and the questions are designed from a visual perspective. This enables a more reliable evaluation of the visual reasoning capabilities of MLLMs.

\section{Related Work}

\begin{figure*}[t!]
\includegraphics[width=\textwidth,trim={0 0 0 0},clip]{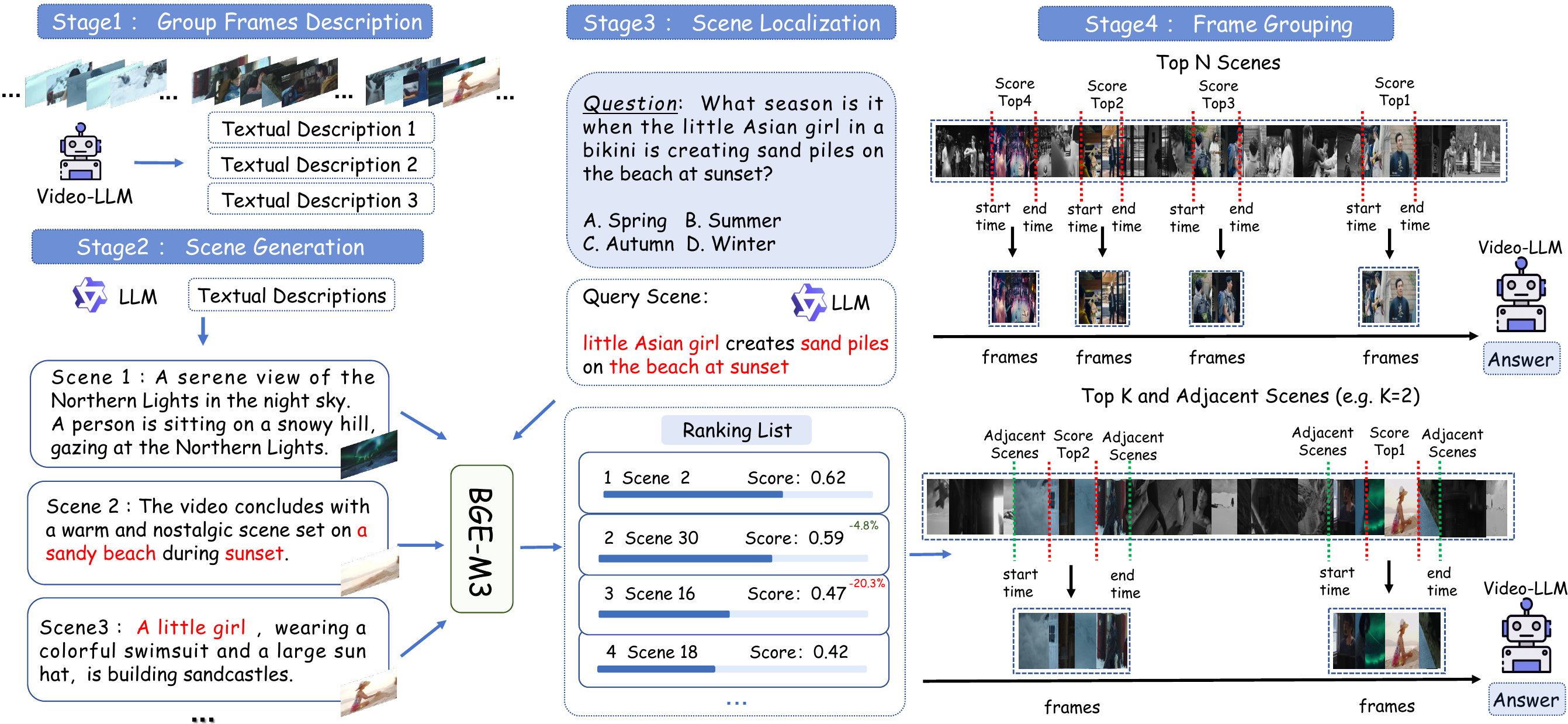}
\caption{
The detailed steps of SLFG. From left to right, the four key stages of the framework are shown: \textbf{Stage 1}: Group Frames and generate visual descriptions with MLLM; \textbf{Stage 2}: Scene Generation with LLM; \textbf{Stage 3}: Scene Localization with embedding models; \textbf{Stage 4}: Group Frames Reorganization for final inference.
} \label{fig3m}
\end{figure*}
\subsection{Long Video Understanding}
Compared to short videos, understanding long videos poses greater challenges, requiring the processing of extensive information over extended periods. 
Some existing methods primarily addressing this challenge through approaches such as compressing visual tokens and enhancing with external modules. For instance, Video-XL \cite{shu2024video} introduces visual summary tokens and dynamic compression strategies to improve the efficiency of long video understanding. MovieChat \cite{song2024moviechat} and MA-LMM \cite{he2024ma} incorporate memory modules with long-term memory banks to achieve accurate predictions for long videos; Video-RAG \cite{luo2024video} utilizes tools such as ASR to enhance external information. Among this, training-based methods like VideoChat-Flash \cite{li2024videochat} and SEAL \cite{wang2024seal} have made progress in this area.
In training-free methods, exploring efficient frame selection has become a new area of research.
Tang \cite{tang2025adaptive} proposes an Adaptive Keyframe Sampling (AKS) algorithm; Video-Tree \cite{wang2024videotree} designs a strategy to adaptively extract key frames relevant to queries from input videos. 
Zhang et al. propose a Q-Frame  \cite{zhang2025q} framework for query-aware frame selection and multi-resolution adaptation.
However, these methods often focus on locating short segments that are directly relevant to the query, overlooking the need for integrated reasoning over semantically coherent scenes

% However, these methods often focus on locating frames directly related to the query, neglecting the need for integrated reasoning over semantically coherent scenes.

\subsection{Long Video Benchmarks}
With the increasing demand for evaluating the long-video understanding capabilities of MLLMs, several benchmarks incorporating longer videos have emerged. LongVideoBench \cite{wu2024longvideobench}, LVBench \cite{wang2024LVBench}, MLVU \cite{zhou2024mlvu}, 
VideoEval-Pro \cite{ma2025videoeval}
MovieChat \cite{song2024moviechat}, MoVQA \cite{tapaswi2016movieqa}, and Video-MME \cite{fu2024video} have started to use longer videos to test the models. Furthermore, research is shifting towards more detailed evaluations. For instance, the Needle-in-a-Haystack task \cite{ye2025re} tests precise specific items, while Grounded QA task like NExT-QA \cite{xiao2024can}  require localizing answers to specific video regions. These developments represent  a growing focus on detail-oriented and evidence-based understanding.

% Among the more mainstream benchmarks, VideoEval-Pro adopts a comprehensive task taxonomy covering four main types and 15 subtypes.
% Video-MME contains 900 videos, and the dataset is also extremely large. In addition to video frames, it also integrates multimodal inputs such as subtitles and audio. However, the drawback is that their average durations (approximately 1017.9 seconds for Video-MME) are still insufficient for a comprehensive understanding of long videos. LVBench has longer video clips, with an average duration of up to 4101 seconds, and the video theme is also very diverse. Multiple themes can relatively comprehensively test the ability of long-term video understanding.
Among them, VideoEval-Pro offers a broad taxonomy with 4 main types and 15 subtypes. Video-MME includes 900 videos with subtitles and audio, but its average video length (about 1018s) is still limited. LVBench provides longer and more diverse videos (average about 4101s), better reflecting long-video scenarios.
Nonetheless, in several current long-video benchmark tests, the task design of some videos is too general or relies on narration, lacking the necessary details. This makes them less suitable for video detail analysis tasks. In contrast, our LVSQA, a dataset designed for SceneQA, focuses on scene-level detail understanding in long videos and provides a more effective evaluation platform for MLLMs.

\section{Methodology}
\subsection{Framework Overview}
In this paper, we present an effective method for long video question answering tasks, which contains four stages. An overview of our proposed framework is shown in Figure~\ref{fig3m}.

In Section 3.2, we will introduce the Group Frames Description in detail, which elaborates on how we sample frames and group them, as well as how MLLMs are used to extract fine-grained visual descriptions from these frame groups. In Section 3.3, we will introduce Scene Generation, explaining how the LLM abstracts scene-level representations from the detailed visual descriptions. In Section 3.4, we will introduce Scene Localization, which details the method of calculating semantic similarity between video scene descriptions and question scenes. And in Section 3.5, we will introduce Group Frames Reorganization, which focuses on how we rearrange and adjust the frame groups based on the similarity scores for the final reasoning process.

\subsection{Group Frames Description}
Building upon the importance of comprehensive visual information, we initiate the process by performing dense frame sampling to ensure consistent and systematic frame extraction from the video. The sampling rate is defined by a fixed time interval \( \Delta t \), such that the \( i \)-th sampled frame is denoted as:
\begin{equation}
f_i = F(t_0 + i \cdot \Delta t), \quad i = 0, 1, 2, \ldots
\end{equation}
where \( F(t) \) represents the video frame at time \( t \), and \( t_0 \) is the initial timestamp.

Then we introduce the grouping granularity N. Each set of $N$ consecutively sampled frames forms a frame group:
\begin{equation}
G_k = \{f_{k}, f_{k+1}, \ldots, f_{k+N-1}\}, \quad k = 0, 1, 2, \ldots
\end{equation}
This grouping strategy allows for a structured organization of the video frames, which aids in downstream processing tasks such as semantic analysis.

For each frame group \( G_k \), we employ MLLMs, denoted as \( \mathcal{M} \), to extract rich semantic content from the visual data. This process is captured as:
% \begin{equation}
% D_k = \mathcal{M}(G_k)
% \end{equation}
\begin{equation}
D_k = \mathcal{M}(G_k) = \{\text{object}_i, \text{action}_j, \text{relation}_{ij}, \ldots\},
\end{equation}
where \( D_k \) represents the descriptive output (in textual form) that conveys the key information extracted from the frames in \( G_k \). The description includes details about the objects, actions, and spatial relationships observed in the frame group.

In this way, the MLLMs process the group of frames \( G_k \) to generate a textual description that captures the essential visual elements, allowing for deeper analysis and generating scene-level representations.

\subsection{Scene Generation}

Subsequently, we employ a LLM (Qwen2.5-7B-Instruct in our experiment) to abstract scene-level representations from the detailed visual descriptions generated in the previous step. Despite the fact that each frame group may potentially contain multiple scenes, the LLM effectively aggregates the information within the descriptions. It identifies and extracts the key scenes, distilling the detailed descriptions into more concise and informative scene-level summaries.

The process of scene generation can be formalized as follows. Given the set of descriptive outputs \( D = \{D_1, D_2, \ldots, D_k\} \) from all frame groups \( G_k \), the LLM aggregates and condenses these descriptions into multiple scene-level representations. Specifically, the LLM identifies and extracts the \( m \)-th scene representation from the descriptions:
\begin{equation}
S_m = \mathcal{L}(D_1, D_2, \ldots, D_k), \quad m = 1, 2, \ldots, M,
\end{equation}
where \( S_m \) represents the \( m \)-th scene-level representation, and \( \mathcal{L} \) is the LLM function that aggregates and abstracts the descriptive outputs \( D_k \) from the frame groups. The function \( \mathcal{L} \) is capable of identifying multiple scenes within the set of descriptions, with each \( S_m \) capturing a distinct scene-level representation. The scene-level representations \( S_m \) are more concise and informative, containing the essential information needed for higher-level tasks. 

The motivation for generating scene-level representations lies in reducing noise and redundancy from low-level descriptions. While \( D_k \) captures fine-grained content, it may contain overlapping or irrelevant details. The LLM aggregates these into concise summaries that better reflect the overall semantic structure. These scene-level summaries \( S_m \) are crucial for linking video content to question intent in subsequent steps. Thus, the process flows in the following structured mapping:
\begin{equation}
f_i \quad \xrightarrow{\text{grouping}} \quad G_k \quad \xrightarrow{\mathcal{M}} \quad D_k \quad \xrightarrow{\mathcal{L}} \quad S_m .
\end{equation}

Each step leads to a more abstract representation, ultimately these scene-level representations serve as a crucial basis for subsequent processing, facilitating precise localization of specific scene information within the video.

\subsection{Scene Localization}
Through the above steps, we have obtained several sets of corresponding scene descriptions. 
Next, we hope to find the most relevant scene to the question, where the correct answer is more likely to be obtained within that group.
an open-source text embedding model (\textbf{BGE-M3} \cite{chen2024bge} in our work) could be used to calculate the semantic similarity between each video scene description and the question scene. Specifically, we encode both the video scene descriptions and the question scene descriptions into dense embeddings, and compute the cosine similarity to measure semantic similarity.

\begin{equation}
\text{Score}_G = \max_{1 \leq i \leq n} \left( \text{Sim}(A_i, B) \right),
\end{equation}

\begin{equation}
\text{Sim}(A_i, B) = \cos \left( \text{emb}(A_i), \text{emb}(B) \right).
\end{equation}
Here, B represents the query scene description, and A$_i$ represents the scene descriptions within the group. We take the highest score from each set as the score for that group, which will be used for ranking.

\subsection{ Group Frames Reorganization}
After obtaining the scores for each group, all groups are sorted in descending order. The sorting results reflect the importance of each scene, where a higher score indicates a higher likelihood that the group contains the relevant segment for answering the question. The next step is to reassemble the scene frames of the top-ranked groups to serve as input for the reasoning module.

A threshold is then applied (\textbf{10\%} in our experiment). If the score difference between two consecutive groups is smaller than the threshold, both groups are considered potential answer segments and are merged. If the score difference exceeds the threshold, the current and all lower-ranked groups are discarded.

Next, we ensure that the total number of frames does not exceed the maximum context window size of the model. If the total number of frames from the selected groups is smaller than the limit, the remaining frames are evenly distributed among the selected groups to extend their context.

Let the number of selected scene groups be \( N \), the number of frames in each group be \( F_i \), the target maximum frame count be \( T \), and the remaining frame budget be R,
the additional number of frames assigned to each group is:
\begin{equation}
\Delta F_i = \left\lfloor \frac{R}{N} \right\rfloor \quad \text{for all} \ i \in [1, N].
\end{equation}
If \( R \mod N \neq 0 \), the remaining frames are distributed based on the time span of each group. Assuming the sampling interval is \( \Delta t \), and the temporal window of each group is \([t_i, t_i + F_i \Delta t]\), the additional frames are sampled both before and after this time range accordingly.

Finally, all the extended scene frames are fed into the model for reasoning to produce the final answer:

\begin{equation}
\text{Output} = \mathcal{M} \left( G_1 + \Delta f,\ G_2 + \Delta f,\ \ldots,\ G_k + \Delta f \right),
\end{equation}
\noindent where $\Delta f = 0$ iff the Group Frames Reorganization strategy is not applied.

% \section{LVSQA : A Dataset Designed for SceneQA }
\section{Dataset }
\subsection{SceneQA Task}
Despite recent advances in video question answering (VideoQA), current task designs still fall short in evaluating a model's ability to comprehend specific scenes within long videos.
% 尽管近年来视频问答取得了显著进展，但现有的任务设计在评估模型对长视频中特定场景的理解能力方面仍显不足。
To bridge this gap, we propose SceneQA, a new task focuseded on scene-localized understanding. In SceneQA, the evaluated  model is required to analyze a specific scene segment and answer questions based on detailed visual cues—such as character actions, object interactions, or scene transitions. This setup strikes a balance between contextual awareness and fine-grained perception, enabling more precise evaluation of video understanding. SceneQA mirrors how humans watch long videos: we attend to key scenes and extract meaningful details. It follows a three-step process—scene localization, detailed perception, and question answering—offering a new framework to advance multimodal long video comprehension.
% To address this gap, we propose SceneQA, a new task that focuses on scene-localized understanding. In SceneQA, the model is expected to focus on a well-defined scene segment and answer questions based on detailed visual cues within that segment—such as character behaviors, object interactions, or scene transitions. This task setup balances the need for contextual awareness with fine-grained visual perception, offering a more targeted way to evaluate real-world video comprehension.
% This task reflects how humans watch long videos: instead of remembering every moment, we focus on key scenes and extract meaningful information from them. SceneQA simulates this cognitive process through three steps: scene localization, detailed perception, and question answering, providing a new perspective for evaluating and advancing multimodal models in long video understanding.

\textbf{SceneQA vs. Traditional VideoQA.}
Traditional VideoQA task is typically based on short clips or isolated segments, where questions can often be answered with limited context. In many cases, models may even rely on subtitles or textual cues to infer answers, rather than truly understanding visual content.  While this setup simplifies reasoning, it fails to assess whether models can understand the temporal and semantic structure of longer narratives. Conversely, SceneQA enables the reasoning over fine-grained visual details within a specific scene after first identifying it.
% 传统的VQA通常基于短视频片段，问题往往可以在有限上下文内完成推理。这种设定简化了理解过程，但不能评估模型是否真正掌握了长视频中的时间结构与语义关联。特别是在处理较长视频内容时这些任务很少涉及场景级推理。

% 这个任务也更贴近人类观看长视频的方式：我们不会逐帧记住每个片段，而是聚焦于关键场景，从中提取有意义的信息。SceneQA就是对这一认知过程的模拟，涵盖三个步骤：场景定位、细节感知与问题回答，为多模态模型在长视频理解任务中的发展与评估提供了新的视角。

% What is largely missing in existing benchmarks is the evaluation of a model's ability to identify a specific scene and reason about fine-grained visual details within it.
%------对比海底捞针and grounded QA-------------

\textbf{SceneQA vs. Needle-in-a-Haystack Task.} Both require locating relevant segments in long videos, but \textit{needle-in-a-haystack} task focus on short, isolated events with precise localization. In contrast, SceneQA targets semantically coherent scenes, demanding integrated reasoning over fine-grained details within them.

\textbf{SceneQA vs. Grounded QA. } Grounded QA (e.g., NExT-GQA \cite{xiao2024can}) links answers to supporting temporal segments, but SceneQA centers on entire scenes as the unit of understanding. It requires distinguishing scene boundaries in long videos and synthesizing scene-wide information for reasoning, beyond GQA's focus on answer-segment consistency.

\begin{figure}[t]
\centering
\includegraphics[width=0.48\textwidth]{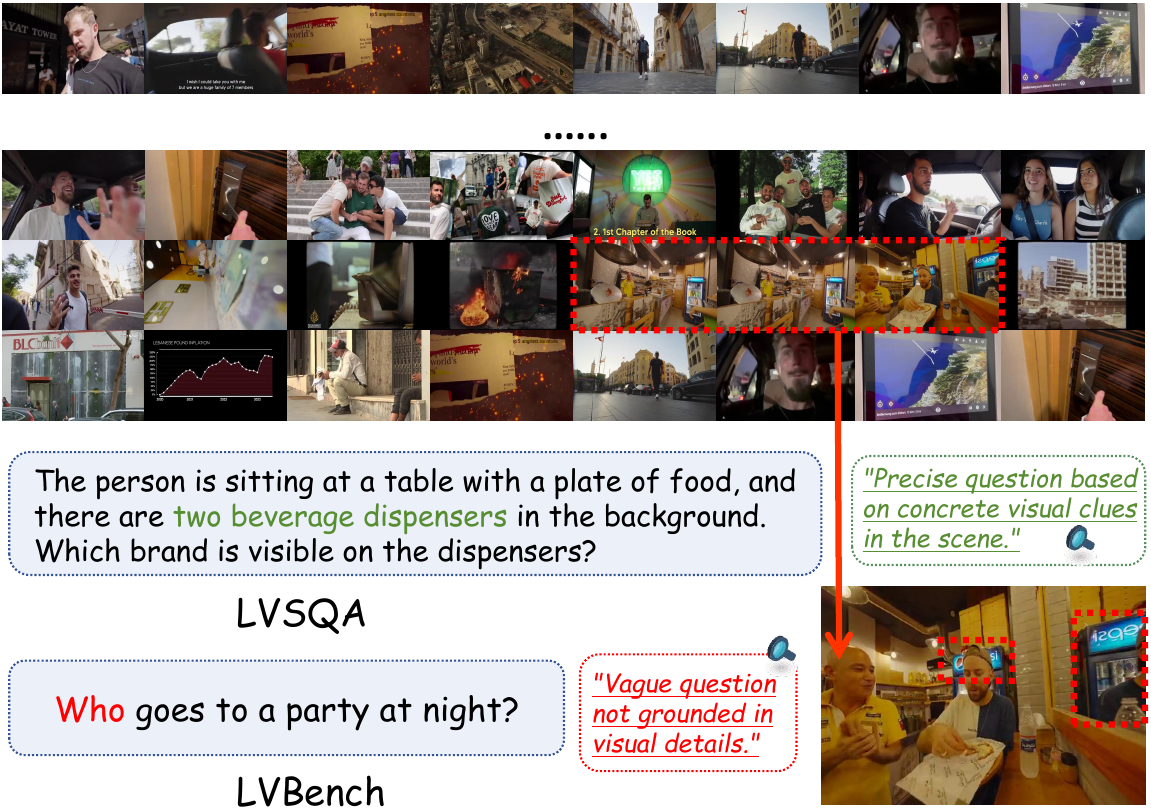}
\caption{A representative example from  LVSQA dataset, which focuses on scene-specific questions that demand detailed visual perception, unlike the general questions in benchmarks like LVBench.} \label{case}
\end{figure}

% \begin{figure}[t]
% \centering
% \includegraphics[width=0.48\textwidth]{figures/lvsqaPipeline.pdf}
% \caption{The five-step pipeline for constructing the LVSQA dataset. These initial pairs then undergo a rigorous, multi-round human refinement and expert filtering process to ensure high quality.} \label{fig5}
% \end{figure}

\begin{table*}
\centering
\small
\setlength{\tabcolsep}{3pt}
\begin{tabular}{c c c c c c c c }
\toprule
\textbf{Model} & \textbf{Size} & \textbf{Type} & \textbf{Frames} &  \textbf{VideoEval-Pro}  & \textbf{LVBench} & \textbf{VideoMME$_{wo~~long}$ } & \textbf{LVSQA}  \\
\midrule
\multicolumn{8}{c}{\emph{\textbf{General MLLMs}}} \\
\midrule
% Oryx‑1.5\cite{ye2024mplug}& 7B & Training-based & 64  & - & - & - & -\\
Video-LLaVA \cite{lin2023video}& 7B & Training-based & 8  & 27.7 & - & 36.2 & 43.6\\
mPLUG-Owl3 \cite{ye2024mplug}& 7B & Training-based & 64  & - & 43.5 & 50.1 & 55.6\\
Qwen2.5-VL \cite{bai2025qwen2} & 7B & Training-based & 64  & 46.9 & 45.3 & 51.0 & 56.8 \\
InternVL3 \cite{chen2024internvl} & 8B & Training-based & 64 & - & - & 48.4 & 55.8 \\
Videollama3 \cite{zhang2025videollama} & 7B & Training-based & 64  & - & 45.3 & 40.6* & 50.8\\
LLaVA-OneVision \cite{li2024llava} & 7B & Training-based & 32  & 40.1* & 38.7 & 43.6* & 56.2\\
LLaVA-Video \cite{zhang2024video} & 7B & Training-based & 64  & 45.8* & 41.8 & 47.6* & 59.8\\
\midrule
\multicolumn{8}{c}{\emph{\textbf{Long Video MLLMs}}} \\
\midrule
Video-CCAM \cite{fei2024video}& 9B & Training-based & 96  & 34.9* & - & 39.6 & 51.2\\
Chat-UniVi-V1.5 \cite{jin2024chat} & 7B & Training-based & 128  & 34.5* & - & 35.8 & 42.8\\
Video-XL \cite{shu2024video} & 7B & Training-based  & 512  & 38.2 & -  & 49.2 & - \\
Kangaroo \cite{ren2025vamba} & 8B & Training-based & 64  & - & 39.4 & 46.7 & 48.8\\
Vamba \cite{ren2025vamba} & 10B & Training-based & 512  & 45.7 & 42.1 & 46.3* & 59.6\\
VideoChat-Flash \cite{li2024videochat} & 7B & Training-based & 512 & 51.2 & 48.2 & 44.1* & 60.1\\
QuoTA \cite{luo2025quota} & 7B & Training-free & 64  & 38.3* & - & 55.7 & 52.8\\
\midrule
\textbf{LLaVA-OneVision + SLFG} & 7B & Training-free & 32  & \textbf{42.6}~\textcolor{red}{(2.5\textuparrow~)} & \textbf{42.5}~\textcolor{red}{(3.8\textuparrow~)} & \textbf{46.9}~\textcolor{red}{(3.3\textuparrow~)} & \textbf{61.0}~\textcolor{red}{(4.4\textuparrow~)}\\
\textbf{LLaVA-Video + SLFG} & 7B & Training-free & 64  & \textbf{48.7}~\textcolor{red}{(2.9\textuparrow~)} & \textbf{45.3}~\textcolor{red}{(3.5\textuparrow~)} & \textbf{49.5}~\textcolor{red}{(1.9\textuparrow~)} & \textbf{63.4}~\textcolor{red}{(3.6\textuparrow~)}\\
\bottomrule
\end{tabular}
\caption{Performance comparison of various MLLMs on benchmarks, including experiments on our proposed LVSQA task and the SLFG method. * indicates the result we reproduced.}
\label{main}
\end{table*}

\subsection{LVSQA Dataset }
To support the SceneQA, we construct \textbf{LVSQA (Long Video Scene-level Question Answering)}, a dataset specifically designed to evaluate fine-grained visual understanding within localized scenes in long videos. 
% Unlike traditional VideoQA task, which often requires models to retrieve information from the entire video, LVSQA constrains the context to a clearly defined scene segment, focusing instead on fine-grained and localized comprehension.

\subsubsection{Core Question Types.}
We categorize the core question types in LVSQA into two main categories:

\begin{itemize}
  \item \textbf{Scene-based Detail Recognition.} This type of question focuses on fine-grained visual details within a specific scene, such as character behaviors, object states, short-term visual changes, or subtle interactions. Solving these questions requires the model to accurately recognize and interpret local visual cues at the frame or subclip level. This type of task tests the model's capacity for precise perception within a constrained temporal window.

  \item \textbf{Scene-based Causal Reasoning.} Building on the previous type, this class of questions requires the model to infer causal relationships within the same scene—understanding what caused a particular action or what consequence followed a specific event. For example, the model may need to infer the motivation behind a character's action, or identify the result of an interaction within the scene. Compared to detail recognition, causal reasoning tasks emphasize the modeling of dynamic event structures and demand deeper scene-level inference.
  
\end{itemize}
Together, these two types define the core challenge of LVSQA: models must not only perceive localized visual information with high quality, but also perform meaningful reasoning under restricted contextual scope.
% This task design reflects a more human-like understanding of long videos, where viewers naturally focus on key scenes and make inferences based on localized content rather than entire narratives.
% 这两种类型共同定义了SceneQA的核心挑战：模型不仅要高质量地感知局部视觉信息，还要在有限的上下文范围内进行有意义的推理。这个任务设计反映了对长视频的更人性化的理解，观众自然会关注关键场景，并根据自然场景的内容而不是整个叙事进行推断。

\subsubsection{Construction Pipeline.}
 The construction of LVSQA follows a rigorous multi-stage pipeline, combining automated generation and large-scale human refinement to ensure the quality, validity, and task alignment of the resulting question-answer pairs. 
 
 % An overview of the pipeline is shown in Figure \ref{fig5}, and the process is elaborated below.

Our dataset construction follows a meticulous five-stage pipeline. \ding{182}  \textbf{\textit{Video Segmentation}}: We begin by selecting 100 long-form videos from the LVBench dataset, each exceeding 30 minutes, which are then uniformly segmented into clips and manually filtered to remove corrupted or irrelevant content. \ding{183} \textbf{\textit{Visual Description Generation}}: For each retained clip, we apply an MLLM to generate fine-grained visual descriptions capturing objects, actions, attributes, and spatial relationships, providing a rich semantic foundation. \ding{184} \textbf{\textit{QA Generation}}: Based on these descriptions, a language model is prompted to generate an initial version of question-answer pairs (QA v1) that target specific visual details for scene localization. \ding{185} \textbf{\textit{Multi-Round Human Refinement}}: This QA v1 set then undergoes a multi-round, double-blind human refinement process, involving over 200 hours of labor to improve clarity, logical consistency, and ensure the pairs are strictly grounded in visual content, yielding a higher-quality QA v2 set. \ding{186} \textbf{\textit{Expert Filtering}}: In the final stage, domain experts perform a rigorous cross-validation of each QA v2 pair against its video segment, a final review requiring over 300 additional hours to ensure every question is fully answerable, coherent, and unambiguous, resulting in the selection of 500 high-quality SceneQA-style pairs for the final LVSQA dataset.

\section{Experimental Evaluation}
\subsection{Experimental Setup and Details}
To evaluate the effectiveness of our proposed SLFG  method, we conduct experiments on multiple long video understanding benchmarks, including VideoEval-Pro, LVBench, VideoMME$_{wo sub~~long}$. These benchmarks cover diverse capabilities, such as multi-modal understanding, video question answering, and fine-grained temporal reasoning. We have also evaluated mainstream MLLMs on our LVSQA task.

SLFG is implemented as a modular component compatible with various open-source MLLMs. We choose LLaVA-OneVision and LLaVA-Video as base models.
The maximum input window is set to 32 and 64 frames espectively, in line with the original experiment \cite{li2024llava}.

We first perform dense sampling at 10-second intervals. Grouping is done with a granularity of N=16. For each group, detailed visual descriptions are generated using MLLM, which are then abstracted into scene representations by Qwen2.5-7B-Instruct \cite{yang2024qwen2}. A 10\% threshold is applied for the recombination strategy. 

To ensure a fair comparison with our method, we do not adopt the lmms-eval \cite{zhang2024lmms} framework. Instead, we implement our evaluation scripts directly based on the provided inference code, which may lead to slightly lower reproducibility results.

All experiments are conducted on 80G A800 GPUs.

\subsection{Experimental Results}
Table~\ref{main} presents the quantitative results on popular video benchmarks. 
On VideoEval-Pro, Our method improves the performance of LLaVA-Video from 40.1\% to 42.6\%, and LLaVA-OneVision from 45.8\% to 48.7\%, demonstrating its effectiveness in boosting general-purpose MLLMs.
On the VideoMME$_{w/o~~long}$ benchmark, our method elevates the performance of LLaVA-OneVision from 43.6\% to 46.9\%, and LLaVA-Video from 47.6\% to 49.5\%.
% When evaluated on the LVSQA benchmark, which emphasizes fine-grained tracking of visual details, our method shows notable improvements. Specifically, LLaVA-OneVision's performance increases from 48.8\% to 53.2\%, a gain of 4.4 percentage points. Similarly, LLaVA-Video improves from 52.2\% to 57.2\%, with an increase of 5.0 percentage points. These findings highlight the advantage of our approach in handling visual detail reasoning tasks effectively.

% On LVSQA task, for Long Video MLLMs, we take 256 frames as input. For General MLLMs, we take 64 frames as input.
We evaluate the LVSQA task across both Long Video MLLMs and General MLLMs. According to the result, the best-performing baselines on the LVSQA benchmark include VideoChat-Flash, with a score of 60.1\%, LLaVA-Video, at 59.8\%, and Vamba, with 59.6\%. In stark contrast, Chat-UniVi-V1.5 and Kangaroo achieve much lower scores, at 42.8\% and 48.8\%. 

This performance gap indicates that the question design in LVSQA effectively differentiates models in their ability to capture and track fine-grained information within long videos, demonstrating the benchmark's well-grounded evaluation dimensions and strong discriminative power. It not only validates the practical utility of the LVSQA task for long video understanding but also underscores the critical importance of long-range temporal modeling and scene-level abstraction in enhancing model performance on complex video comprehension tasks.
We will incorporate more models into the LVSQA task evaluation. To avoid accuracy discrepancies caused by our own reproduction, we will also maintain a leaderboard and welcome the community to submit their own answers.

\subsection{Exploratory Experiments}
\subsubsection{Comparison of Different Retrieval Methods.}
Our experimental results reveal a clear distinction between the paradigms for long video understanding. This comparison centers on two baselines: the keyframe retrieval, represented by Adaptive Keyframe Sampling (AKS), and the query-oriented token allocation, represented by QuoTA.
AKS uses a vision-language model like CLIP to find frames relevant to a query. QuoTA is a training-free method that uses Chain-of-Thought to understand a query and then allocates more of a fixed token budget to the most relevant frames.

Our analysis leads to a core conclusion: retrieval-based methods like AKS are effective for ``finding things", while our scene-centric SLFG is essential for ``understanding the story". For simple fact-finding tasks (like Local Perception in VideoEval-Pro), AKS performs competitively by directly matching text to visual features. However, for complex narrative reasoning (like Holistic Reasoning), this focus on isolated frames is a fundamental limitation.

This is where the superiority of our scene-level understanding becomes clear. SLFG treats a video not as a collection of frames, but as a series of coherent scenes. This preserves the context and logic needed to understand plot and causality. We believe the core challenge of long video understanding is this narrative comprehension. Therefore, SLFG's superior performance in reasoning tasks, proves its greater suitability for the field's fundamental challenges.

\begin{table}[ht]
\small
\setlength{\tabcolsep}{3pt}
\renewcommand{\arraystretch}{1.3}
\centering
\label{tab:videoeval-pro-hline}
\begin{tabular}{cccc}
\hline
\textbf{Category} & \textbf{QuoTA} & \textbf{AKS} & \textbf{SLFG (Ours)} \\
\hline
\textbf{Local Perception} & 43.5 & \textbf{55.3} & 54.6 \\
\textbf{Local Reasoning} & 36.7 & 44.2 & \textbf{48.3} \\
\textbf{Hololic Perception} & 33.0 & 28.9 & \textbf{33.2} \\
\textbf{Holistic Reasoning} & 22.6 & 36.4 & \textbf{39.6} \\
\hline
\textbf{Overall Average} & 38.3 & 47.7 & \textbf{48.7} \\
\hline
\end{tabular}
\caption{Macro-level performance comparison on the four main categories of the VideoEval-Pro. All scores are reported as accuracy (\%).}
\end{table}
% The categories are LP (Local Perception), LR (Local Reasoning), HP (Hololic Perception), and HR (Holistic Reasoning).
\subsubsection{Time Efficiency of SLFG.}

In terms of efficiency, our method exhibits significant advantages as the number of questions increases for the same video. This is because the Group Frames Description and Scene Generation stages are executed only once and reused across all questions associated with a single video. By avoiding redundant computations for each new query, our framework greatly reduces overall processing time.

We conduct experiments on a 30-minute video selected from the LVSQA, with the grouping granularity fixed at 16 frames. Since most videos in the dataset contain only five questions, we simulate denser interaction scenarios by repeating the inference process twice, effectively producing ten questions per video.We compare our method against three configurations: the original LLaVA-Video baseline, LLaVA-Video with AKS, and LLaVA-Video with QuoTA. We record the average inference time per question under each setting.

\begin{figure}[htbp]
\centering
\includegraphics[width=0.44\textwidth]{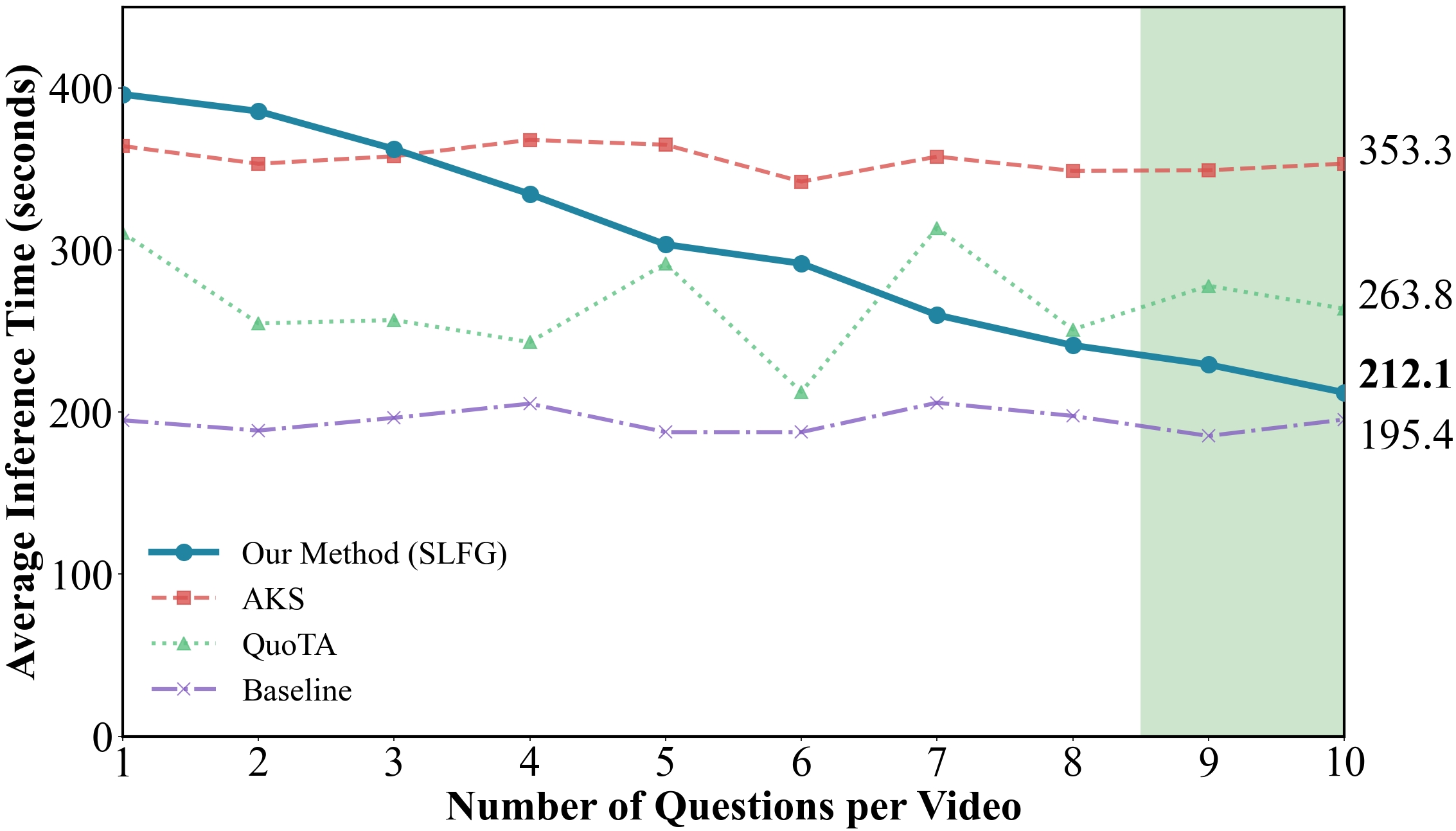}
\caption{Efficiency improvement of our method as the number of questions increases for the same video.}
\label{fig6}
\end{figure}

Experimental results show that when a video contains approximately 10 questions, the average inference time per question using our method approaches that of direct inference without any preprocessing. As the number of questions increases further, the average time per question continues to decrease, demonstrating the scalability and efficiency of our approach for videos with a large number of questions. As shown in Figure \ref{fig6}, the efficiency of our method improves as the number of questions increases.

\subsubsection{Effects of Group Frames Reorganization Strategy.}

To evaluate the effectiveness of our proposed group frames reorganization strategy, we design an ablation study with three experimental settings:

We compare several scene selection strategies: \ding{182} \textbf{\textit{M1 (Top-1 Only):}} Selects only the highest-scoring group for reasoning without any merging or extension, representing the simplest approach.\ding{183} \textbf{\textit{M2 (Top-N without Merging):}} Selects the top-N highest-scoring groups without applying any merging or temporal padding, which may introduce redundant information.\ding{184} \textbf{\textit{M3–M5 (Dynamic Reorganization):}} Our proposed strategy that merges similar groups and extends the time window to include surrounding frames. Specifically, M3 uses a 5\% threshold, M4 uses a 10\% threshold, and M5 uses a 20\% threshold for merging.

% \begin{itemize}[left=0pt, itemindent=0pt, wide=0pt] 
%     \item M1 (Top-1 Only): Only the highest-scoring group is selected for reasoning without any merging or extension. This represents the simplest approach.
%     \item M2 (Top-N without Merging): The top $N$ highest-scoring groups are selected, but no threshold-based merging or temporal padding is applied. This may lead to redundant information.
%     \item M3 (Dynamic Reorganization with 5\% Threshold): Our strategy with a 5\% threshold for merging similar groups and extending the time window to include surrounding frames.
%     \item M4 (Dynamic Reorganization with 10\% Threshold): Our strategy with a 10\% threshold for merging similar groups and extending the time window to include surrounding frames.
%     \item M5 (Dynamic Reorganization with 20\% Threshold): Our strategy with a 20\% threshold for merging similar groups and extending the time window to include surrounding frames.
% \end{itemize}

We conduct experiments using LLaVA-Video on the LVSQA dataset. As shown in Table~\ref{tab:1}, the strategy with varying thresholds shows that the 10\% threshold results in the best performance, followed by the 20\% threshold. The 5\% threshold, while still effective, is less efficient in preserving relevant semantic information. Our method achieves superior performance, the best-performing threshold results outperforming M1 by 2.2 percentage points and M2 by 0.4. 

\begin{table}[H]
\small
\centering
\renewcommand{\arraystretch}{1.3}
\begin{tabular}{c|c|c}
\hline
\textbf{Strategy} & \textbf{Acc (\%)} & \textbf{Over M4 } \\
\hline
 \textbf{M1 (Top-1 Only)} & 61.2 & -2.2\% \\
 \textbf{M2 (Top-N)} & 63.0 & -0.4\% \\
\hline
\multicolumn{3}{c}{\textit{ \textbf{Our Dynamic Reorganization}}} \\
\hline
 \textbf{M3 (5\% Threshold)}  & 63.2 & -0.2\% \\
 \textbf{M4 (10\% Threshold)} & \textbf{63.4} & -  \\
 \textbf{M5 (20\% Threshold)} & 63.0& -0.4\% \\
\hline
\end{tabular}
\caption{Ablation study results for group frames reorganization strategy and the threshold selection.}
\label{tab:1}
\end{table}
\section{Conclusion}
In this paper, we propose a new scenario SceneQA under the video question answering task, which focuses on MLLMs’ detail perception in long videos at the scene level. Based on this, we introduce the SLFG method, whose core idea is to leverage scene-level semantic information and mimic human reasoning patterns to reduce information redundancy. This opens up a new direction for exploration in the field of video understanding.
To more fairly evaluate the detail perception capabilities of multimodal large language models in long-video scenarios, we further augment LVBench with new scene-level evaluation QA pairs, termed LVSQA. We will maintain a leaderboard to provide a new dataset for comparing various strategies in long-video tasks.

\bibliography{aaai2026}

\end{document}